\documentclass[runningheads]{llncs}

\usepackage[T1]{fontenc}

\usepackage{graphicx}
\usepackage{authblk}
\usepackage{blindtext}

\begin{document}

\title{RepVGG-GELAN: Enhanced GELAN with VGG-STYLE ConvNets for  Brain 
Tumour Detection}

\author{Thennarasi Balakrishnan\inst{1},
Sandeep Singh Sengar\inst{2}}
\authorrunning{T. Balakrishnan, S.S. Sengar}
%
\institute{School of Technologies, Cardiff Metropolitan University, Cardiff, United Kingdom.
\\
\email{\inst{1}thennarasibalakrishnan@gmail.com, \inst{2}SSSengar@cardiffmet.ac.uk}
}

\maketitle              
\begin{abstract}

Object detection algorithms particularly those based on \linebreak YOLO have demonstrated remarkable efficiency in balancing speed and accuracy. However, their application in brain tumour detection remains underexplored. This study proposes RepVGG-GELAN, a novel YOLO architecture enhanced with RepVGG, a reparameterized convolutional approach for object detection tasks particularly focusing on brain tumour detection within medical images. RepVGG-GELAN leverages the RepVGG architecture to improve both speed and accuracy in detecting brain tumours. Integrating RepVGG into the YOLO framework aims to achieve a balance between computational efficiency and detection performance. This study includes a spatial pyramid pooling-based Generalized Efficient Layer Aggregation Network (GELAN) architecture which further enhances the capability of RepVGG. Experimental evaluation conducted on a brain tumour dataset demonstrates the effectiveness of RepVGG-GELAN surpassing existing RCS-YOLO in terms of precision and speed. Specifically, RepVGG-GELAN achieves an increased precision of 4.91\% and an increased AP50 of 2.54\% over the latest existing approach while operating at 240.7 GFLOPs. The proposed RepVGG-GELAN with GELAN architecture presents promising results establishing itself as a state-of-the-art solution for accurate and efficient brain tumour detection in medical images. The implementation code is publicly available at {https://github.com/ThensiB/RepVGG-GELAN}.


\end{abstract}
\section{INTRODUCTION}

With the high rates of sickness and mortality, brain tumors represent a significant global health concern. Brain tumour identification challenges can be effectively resolved with automated detection techniques that make use of state-of-the-art technologies like deep learning algorithms. Incorporating automated detection into medical processes has the potential to significantly enhance patient outcomes and medical services by revolutionizing the management of brain tumours, especially as technology develops. The cutting-edge object detection method YOLO divides the input image into a grid while estimating class probabilities and bounding boxes for each grid cell~\cite{ref_url1}. The application of YOLO in brain tumour detection has significant promise for improving the accuracy, efficiency, and scalability of neuroimaging diagnosis approaches~\cite{ref_url2}.
Convolutional Neural Networks (CNN) are widely employed as the primary component of YOLO object identification methods. CNN provides the feature extraction capability needed to recognize objects in an image. Because CNN employs several convolutional and pooling layers to create a hierarchical feature model from raw visual data, they can catch complex patterns and structures that could be indicative of various health problems~\cite{ref_url3}. The VGG deep convolutional neural network architecture was proposed in response to the need for more sophisticated networks with deeper learnable parameters. Further, inside the network max-pooling layers come after several convolutional layers. Their dense hierarchical structure allows for exact detection and diagnosis, consequently, they can recognize complex patterns and features from medical images. VGG architectures are excellent at extracting complicated features and fine-grained information from medical images making them suitable for tasks requiring high-resolution analysis~\cite{ref_url4}. The objective of ShuffleNet, a more contemporary convolutional neural network architecture is to maximize accuracy while maintaining competitive computational efficiency~\cite{ref_url5}. This introduces the concept of group convolutions and channel shuffling which drastically reduces processing costs while promoting efficient data flow between channels. ShuffleNet architecture provides a reasonable balance between accuracy and efficiency making them ideal for low-power device deployment and real-time medical imaging applications~\cite{ref_url1,ref_url6}.
The primary concept behind real-time object detection in the YOLO architecture was to use spatially separated bounding boxes with matching class probabilities as a regression task. To boost detection accuracy and speed, YOLOv4 incorporated several enhancements. Using cross-stage partial connections it included the CSPDarknet53 backbone which improved gradient propagation and information flow. With the implementation of AF (Anchor-Free) detection in YOLOv7, anchor boxes were rendered obsolete. This enhanced adaptability and streamlined layout. Additionally, YOLOv7 incorporated Dynamic Convolution which enhanced the model's ability to extract contextual data by dynamically modifying the receptive field in response to input feature values. YOLOv9 is an evolution of the YOLO object detection method which is admired for its real-time detection capabilities. By combining PGI and GELAN, YOLOv9 introduces numerous architectural advances and training methods that result in improved accuracy and performance. YOLOv9 is built upon the framework of YOLOv7 and Dynamic YOLOv7 and includes RepConv and GELAN with CSPNet blocks. Simplified down-sampling modules and optimised prediction heads without anchors are achieved. The auxiliary loss part of PGI comes after YOLOv7's auxiliary head setup~\cite{ref_url1,ref_url7}.
Based on the Reparametrized Convolution and Generalized Efficient Layer Aggregation Network (GELAN), this study suggests a unique YOLO model called RepVGG-GELAN to develop a high-accuracy object detector for medical images. The purpose of RepVGG-GELAN is to combine the advantages of GELAN and RepVGG models for object detection applications. 

The subsequent sections of this paper are structured as follows, Section 2 provides an overview of related studies. Section 3 delves into the detailed explanation of the proposed RepVGG-GELAN model. Section 4 presents the obtained results with the implementation and analysis of the outcomes. Finally, the Section 5 includes the conclusion part.

\section{LITERATURE REVIEW}

\subsection{RepVGG/RepConv ShuffleNet}

Inspired by ShuffleNet, the RepVGG/RepConv ShuffleNet (RCS) approach employs structural reparameterized convolutions to enhance feature extraction while reducing computational expenses~\cite{ref_url2}. 

\begin{figure}
\includegraphics[width=\textwidth]{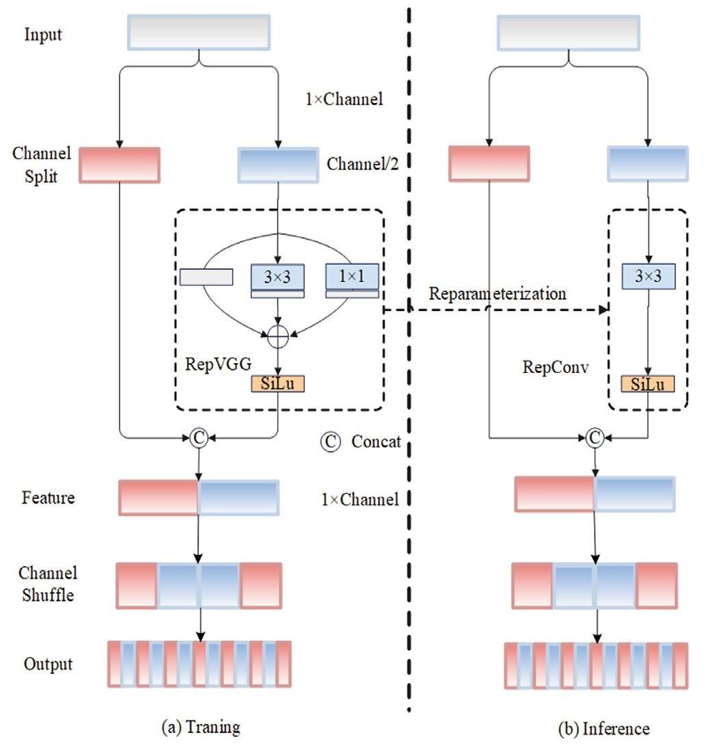}
\caption{The RepVGG architecture. (a) during the training phase, (b) during the inference phase (or deployed).\\
Source: (RCS-YOLO: A Fast and High-Accuracy Object Detector for Brain Tumor Detection, 2023)
} \label{fig1}
\end{figure}

As shown in Fig.~\ref{fig1}, using a channel split operator the input tensor which has dimensions of C x H x W for channels, height, and width, respectively is split into two tensors of equal dimensions. During training, each tensor is processed using multiple convolutional blocks. This consists of the 3x3 convolution, 1x1 convolution, and identity branch. Data from both branches is combined upon processing by concatenating the tensors channel-wise. A single tensor is subjected to numerous convolutional steps to gather a large amount of feature data for training. The model can adapt to various object detection circumstances including those with complicated backgrounds or occlusions because of this diversity in feature learning. A single 3x3 RepConv contains all of the training methods that utilise structural reparameterization during inference. This optimisation allows the model to conclude quicker and with reduced memory use which makes it useful for circumstances requiring immediate actions or low resource availability. By improving information fusion between the two tensors the channel shuffle operator improves the efficiency of feature representation. The ability of the channel shuffle operator to properly combine feature information from multiple convolution groups enhances the model's feature extraction performance. The computational cost can be reduced by a factor of 1/g with the channel shuffle operator, where 'g' represents the total number of groups in the aggregated convolutions. Compared to conventional 3x3 convolutions RCS maintains inter-channel data transfer while reducing the computing cost during inference by a factor of two as shown in Fig.~\ref{fig1}(b). Consequently, RCS can effectively collect contextual information and spatial connections needed for precise object detection. Using cutting-edge methods including reparameterization, shuffle, and channel splitting, RCS produces outstanding feature representation. RCS provides fast and effective inference which could reduce computational complexity and memory utilisation for real-time applications. Because the structure allows for robust representation learning from input data during training it improves the model's capacity to recognise complex patterns~\cite{ref_url2}.
RCS is an excellent option for practical object detection applications since it strikes a compromise between computational efficiency and feature representation through the use of channel split, shuffle, and reparameterization algorithms. Its versatility in learning different properties, memory optimisation, and effective inference makes it an essential tool in computer vision~\cite{ref_url2}.

\subsection{Generalized Efficient Layer Aggregation Network (GELAN)}

GELAN provides a task-dependent advanced method. The input tensor represents an array of images fed into the GELAN object detection model. All images in batches have predetermined channels, heights, and widths~\cite{ref_url8}. 

\begin{figure}
\includegraphics[width=\textwidth]{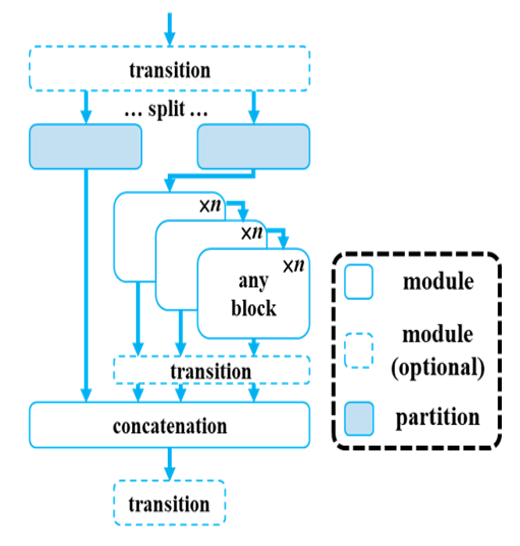}
\caption{Generalized Efficient Layer Aggregation Network Architecture\\ Source: (YOLOv9: Learning What You Want to Learn Using Programmable Gradient Information, 2024)} \label{fig2}
\end{figure}

Fig. 2 explains the architecture of GELAN, a lightweight network architecture based on gradient path planning. The input tensor undergoes many convolutional operations in the backbone layers. In certain circumstances, the number of channels can be increased while the spatial dimensions (height and width) of the feature maps decrease. With these layers' capacity to extract features at different levels of abstraction, the model can obtain both high-level semantic information and fundamental details that are crucial for object detection. As the input tensor passes through the backbone levels hierarchical feature extraction is applied to it. Every layer captures features at various levels of abstraction. After receiving the input tensor SPP (Spatial Pyramid Pooling) blocks process the feature maps to extract multi-scale features. Through adaptive pooling methods, SPP blocks extract features at different spatial resolutions, making them more resilient to object size changes and occlusions. Further, the RepNCSPELAN4 (Rep-Net with Cross-Stage Partial CSP and ELAN) block enhances and refines the feature representations by processing the input tensor. RepNCSPELAN4 blocks integrate convolutional layers to train discriminative features. These architectural components are designed to manage input properties efficiently while preserving the spatial and semantic information required for accurate object detection. Upsampling techniques are applied at various network layers to enhance the spatial resolution of feature maps. Multi-scale data integration is facilitated by concatenating the upsampled feature maps with feature maps from earlier layers. This makes it feasible for the model to preserve fine-grained characteristics and spatial connections which leads to improved object localization and recognition. After the feature maps are evaluated the detection head generates predictions for object detection. The detection head generates bounding boxes, class probabilities, and other relevant information for each object detected in the receiving images. The GELAN's Detect module receives feature maps from different detection layers and uses them to provide class confidence ratings and bounding box predictions. To generate predictions the convolutional layers in the forward pass are applied to the input feature maps. Class confidence ratings and predictions for bounding box regression are computed using the output from these layers. During inference, the module dynamically computes strides and anchor boxes based on the structure of the input feature map. These anchor boxes are employed to decode the bounding box predictions. The anchor boxes and strides are dynamically changed if the input changes. GELAN's modular and flexible architecture allows it to be readily tailored to a wide range of datasets and object detection applications. The model can detect a large variety of object attributes and spatial correlations due to its reparameterized convolutional blocks, spatial pyramid pooling, and hierarchical structure. Its adaptable structure and inference speed make it ideal for practical implementation in a range of computer vision applications from object detection in self-driving vehicles and medical imaging to pedestrian detection in surveillance systems~\cite{ref_url8,ref_url9,ref_url10}.

\section{METHODOLOGY}

The study proposes a novel YOLO model, RepVGG-GELAN shown in Fig. 3, based on the RepVGG/RepConv to create a high-accuracy object detector for medical images. RepVGG-GELAN is designed to leverage the strengths of both RepVGG and GELAN models for object detection tasks. 

\begin{figure}
\includegraphics[width=\textwidth]{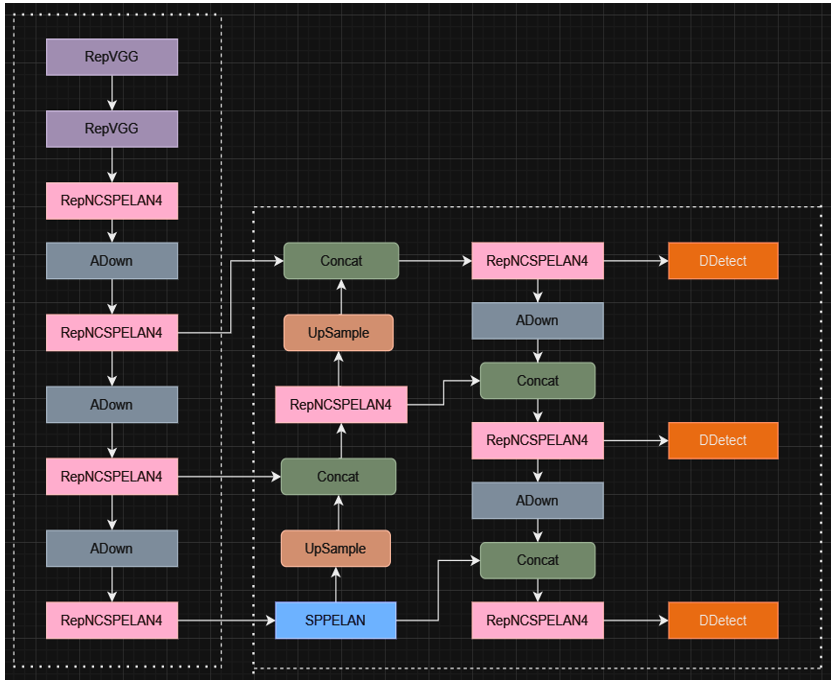}
\caption{Proposed RepVGG-GELAN. The architecture of RepVGG-GELAN is mainly comprised of RepVGG and RepNCSPELAN4.} \label{fig3}
\end{figure}

RepVGG (Reparametrized VGG) is a simplified convolutional neural network (CNN) architecture that combines depthwise separable convolutions and residual connections. RepVGG blocks provide the starting point of RepVGG-GELAN and are renowned for their ease of use and efficacy in feature extraction. ReLU and identity mappings are combined in these blocks to replace conventional convolutional layers improving training stability and performance~\cite{ref_url2,ref_url5,ref_url13}.  

RepNCSPELAN4 (Rep-Net with Cross-Stage Partial CSP and ELAN) is a block architecture that combines Cross-Stage Partial (CSP) connections and ELAN for feature enhancement. It splits the input into two parts processes each part separately with multiple RepNCSP blocks and then concatenates the outputs before passing them through the final convolutional layer. CSP connections facilitate information flow between different stages of the network. ELAN enhances feature representation through attention mechanisms. RepNCSPELAN4 enhances feature representation by combining efficient feature extraction with attention mechanisms.
It enables the network to capture and emphasize important features relevant to object detection tasks~\cite{ref_url8,ref_url14}.

ADown module represents an asymmetric downsampling block. It takes an input tensor x, performs average pooling on one half and max pooling on the other half, applies convolutional operations to each pooled tensor and concatenates the results before returning. This block is useful for downsampling feature maps with different operations on each half~\cite{ref_url8,ref_url14}.

Spatial Pyramid Pooling with ELAN consists of a series of convolutions followed by spatial pyramid pooling (SPP) operations where the feature maps are divided into regions of different sizes and the features are pooled from each region separately. The pooled features are then concatenated and processed through another convolutional layer. This block is useful for capturing multi-scale information from the input feature maps~\cite{ref_url8,ref_url14}.

Upsampling and Concatenation operations up-sample feature maps from the backbone and concatenate them with features from previous stages.
They enable multi-scale feature fusion and preserve spatial information~\cite{ref_url8,ref_url14}.

The DDetect block processes input feature maps through convolutional layers to predict bounding box coordinates and class probabilities. It utilizes predefined anchor boxes and strides for inference. The biases in the detection head are initialized based on the nominal class frequency and image size. Bias initialization helps ensure that the detection head starts with reasonable predictions during training~\cite{ref_url8,ref_url14}.

The overall result of RepVGG-GELAN is an effective object detection architecture that combines the efficiency and simplicity of RepVGG with the sophisticated feature aggregation and processing capabilities of GELAN through the combined use of both models' advantages. RepVGG-GELAN seeks to enhance performance, efficiency, and accuracy in difficult detection tasks.

\section{EXPERIMENTS AND RESULTS}

\subsection{Data Collection}
To assess the proposed RepVGG-GELAN model employed the brain tumour detection 2020 dataset (Br35H)~\cite{ref_url11}, which comprises 701 images spread across two folders, labelled `TRAIN' and `VAL'. Each folder has 2 subfolders named labels and images, where the labels of each image are stored in a text file in the labels folder. The .txt format annotations in the folder dataset-Br35H are converted from the original JSON format.  Out of these 701 images 500 images were designated as the training set and the remaining 201 images as the testing set. The input image size is set to $640 \times 640$. The label boxes of the brain images were normalized and are in the format class, xcenter, ycenter, width, and height.

\subsection{Implementation details}

The following configurations and setups were employed at Google Colab to develop the RepVGG-GELAN model. Windows 11 operating system is used with the processor for CPU: Intel Iris Xe, Deep Learning Framework: PyTorch 1.9.1, GPU: NVIDIA GeForce RTX 3090 with 24GB memory capacity (available via Google Colab), CUDA Toolkit used through the Google Colab environment. The model was trained for a maximum of 150 epochs with the number of batches set to 8, the resolution of the image as 640 x 640 pixels, Stochastic Gradient Descent (SGD) optimisation at 0.937 with an initial learning Rate of 0.01 and weight decay coefficient 0.0005. Linear warm-up is used in the first three epochs. Weight decay adds a penalty term to the loss function to prevent overfitting by penalising large weights. Mosaic augmentation is applied to every training sample and is turned off for the last 15 epochs. The Colab environment makes it easier to use CUDA for deep learning applications more effectively. 

The model loads the configuration from rcs-gelan-c.yaml file with the number of input channels ('ch') 1 and the number of classes ('nc') 1 [0: brain tumour]. Convolutional layers (Conv2d()) and batch normalisation layers (BatchNorm2d()) are fused to optimize the inference speed. The input feature maps 'x' from each detection layer are concatenated and returned during training. During inference, anchor boxes and strides are computed dynamically if needed. Bounding box predictions and class predictions are extracted from the concatenated output. The IoU (Intersection over Union) is calculated as the ratio of the intersection area to the union area, where the intersection area is calculated as the product of the widths and heights of the intersecting regions after taking the maximum of zero and the differences between the coordinates and the union area is calculated as the sum of the areas of the individual bounding boxes minus the intersection area. Bounding box coordinates are refined using the Down-sampling Feature Localization (DFL) layer (adjusts the centre coordinates and scales the width and height based on learned parameters). The bounding box predictions undergo refinement to improve localization accuracy.
Bounding box coordinates are transformed and descaled based on anchor boxes and strides. Class predictions are passed through a sigmoid activation function. The final output consists of the concatenated predictions, which include the transformed bounding box coordinates and the sigmoid-activated class scores.

\subsection{Evaluation metrics}
To assess the benefits and drawbacks of the model this study employs FLOPs (Floating Point Operations), mAP50 (mean Average Precision at IoU threshold of 0.5), mAP50:95 (mean Average Precision over IoU thresholds from 0.5 to 0.95), precision, and recall as comparative measures of detection impact in this work. The following formulas are used to determine precision and recall with IoU (Intersection over Union) value 0.5: 

\begin{equation}
Precision = TP / (TP + FP)
\end{equation}

\begin{equation}
Recall = TP / (TP + FN )
\end{equation}

where FP stands for negative samples mistakenly recognised as positive samples, FN for positive samples mistakenly identified as negative samples and TP for the number of positive samples correctly identified as positive samples. FLOPs, representing the number of floating point operations required for inference, help in understanding the computational cost associated with running the model on different hardware platforms. mAP@0.5 average precision calculated for brain tumour class at a specific IoU (Intersection over Union) threshold of 0.5, considers the precision-recall trade-off for the class and computes the average precision. mAP@0.5:0.95 a more comprehensive metric that evaluates the model's performance across a range of IoU thresholds provides a broader understanding of how well the model localizes objects at various levels of overlap with ground truth bounding boxes.

\subsection{Results}

\begin{table}
\caption{Quantitative results of RCS-YOLO, YOLOv8 and RepVGG-GELAN. The best results are shown in bold.}\label{tab1}
\begin{tabular}{|l|l|l|l|l|l|}
\hline
  &  Precision & Recall & AP50 & AP50:95 & Parameters
  (in millions)\\
\hline
RCS-YOLO &  0.936 & 0.945 & 0.946 & 0.729 & 45.7\\
YOLOv8 &  0.973 & \textbf{0.909} & 0.957 & \textbf{0.733} & 30.1\\
{\bfseries RepVGG-GELAN} &  {\bfseries0.982} & 0.890 & {\bfseries0.970} & 0.723 & {\bfseries25.4}\\
\hline
\end{tabular}
\end{table}

The RepVGG-GELAN model achieves an exceptional precision score of 0.982 indicating its remarkable ability to correctly identify true positive cases while minimizing false positives. Although the recall score of 0.890 is slightly lower compared to other models, it still demonstrates the model's effectiveness in capturing a substantial proportion of actual positive cases. RepVGG-GELAN achieves an impressive AP50 score of 0.970 surpassing both RCS-YOLO and YOLOv8, demonstrating its effectiveness in accurately localizing objects with sufficient overlap with ground truth bounding boxes. This metric reflects the model's ability to maintain high precision across different IoU thresholds, particularly at the 50\% threshold (commonly used in object detection tasks). With an AP50:95 score of 0.723 RepVGG-GELAN demonstrates consistent performance across a wider range of IoU thresholds. While slightly lower compared to RCS-YOLO and YOLOv8, this metric still reflects the model's robustness in detecting brain tumours across varying levels of overlap between predicted and ground truth bounding boxes. One of the key strengths of RepVGG-GELAN is its efficiency in model size with only 25.4 million parameters. This streamlined architecture ensures computational efficiency without compromising performance making it well-suited for deployments. 

\subsection{Ablation study}

\begin{table}
\caption{Ablation results on the proposed RepVGG-GELAN module. The best results are shown in bold.}\label{tab2}
\begin{tabular}{|l|l|l|l|l|}
\hline
  &  Precision & Recall & AP50 & AP50:95\\
\hline
{\bfseries RepVGG-GELAN (w RepVGG)} &  {\bfseries0.982} & 0.89 & {\bfseries0.97} & {\bfseries0.723}\\
RepVGG-GELAN (w/o RepVGG) & 0.964 & 0.902 & 0.957 & 0.722 \\
\hline
\end{tabular}
\end{table}

RepVGG-GELAN achieves a higher precision of 0.982 compared to GELAN's precision of 0.964. This indicates that RepVGG-GELAN is more accurate in correctly identifying positive detections leading to a reduction in false positives. Moreover, despite having a slightly lower recall of 0.89 compared to GELAN's recall of 0.902, RepVGG-GELAN achieves a higher mAP50 of 0.97 indicating better overall detection performance across different thresholds. By leveraging the strengths of both RepVGG and GELAN architectures RepVGG-GELAN achieves higher precision and overall detection performance, making it a more effective and reliable model.

\section{CONCLUSION}
This study was focused on the development and evaluation of the RepVGG-YOLO model for the detection of brain tumours in medical imaging data. The incorporation of GELAN architecture into RepVGG enhances the model's ability to extract relevant features from medical imaging data leading to improved tumour detection accuracy. The model was designed with a focus on efficiency, leveraging techniques such as structural reparameterization and efficient layer aggregation to optimize resource utilization and inference speed.

\end{document}